\documentclass[letterpaper]{article}
\usepackage[preprint]{aaai2027}
\usepackage[hyphens]{url}
\usepackage{graphicx}
\usepackage{natbib}
\usepackage{caption}
\usepackage{booktabs}
\usepackage{amsmath}
\usepackage{amssymb}
\usepackage{tabularx}
\newcommand{\method}{FreqID-Control}
\title{\method: Frequency Routing and Identity Conditioning for Thermal-to-Visible Face Translation}

\author{
Haojie Li\textsuperscript{1}, Xinlong Sun\textsuperscript{2}, Xiaohaoyang Lei\textsuperscript{3}, Kai Cheng\textsuperscript{3},\\
Yiguo Qiao\textsuperscript{2}, Xu Yang\textsuperscript{2}, Cunjian Chen\textsuperscript{3}
}
\affiliations{
\textsuperscript{1}Monash University, Monash Suzhou Research Institute, Suzhou, China\\
\textsuperscript{2}Southeast University, School of Computer Science and Engineering, PALM Lab, Nanjing, China\\
\textsuperscript{3}Monash University, Faculty of Information Technology, Australia
}

\begin{document}

\maketitle

\begin{abstract}
Thermal-to-visible (T2V) face translation has important applications in nighttime biometric recognition, low-light surveillance, and human interpretation of thermal observations. The task remains challenging because thermal images do not directly record visible appearance, while the cross-spectral cues useful for reconstruction vary across frequency components and semantic roles. Existing spatially conditioned models tend to process these cues through a homogeneous adaptation path and do not explicitly separate local structure from global identity. We propose \method{}, a parameter-efficient framework that organizes thermal guidance according to spectral reliability and semantic role while preserving the pretrained generative prior. It combines frequency-routed low-rank adaptation, which allocates separate transformation subspaces to low-frequency structure and high-frequency detail, with shared cross-spectral identity conditioning, which supplies compact global identity tokens independently of spatial thermal tokens. A controlled analysis on MCXFace and SpeakingFaces provides the empirical basis for this design, revealing stable low-frequency correspondence and edge-localized detail correspondence. Experiments on both datasets demonstrate strong image fidelity and biometric identity preservation, while controlled ablations on MCXFace confirm the complementary contributions of frequency routing and shared identity conditioning.

\end{abstract}

\section{Introduction}

Thermal-to-visible (T2V) face translation supports cross-spectral biometric recognition in nighttime surveillance, low-light security, and search-and-rescue scenarios. When visible-light sensors cannot provide reliable observations, thermal cameras can still capture facial patterns associated with emitted radiation. However, thermal imagery does not record reflected visible appearance and therefore cannot uniquely determine pigmentation, reflectance, or fine visible-spectrum textures. The central objective of T2V face translation is thus not to recover every unobserved appearance attribute, but to preserve the facial structure and identity information shared across spectra while using a visible-domain generative prior to synthesize plausible domain-specific details.

Existing approaches to T2V face translation have improved cross-spectral synthesis through attribute-preserving synthesis \citep{di2018attribute}, identity-spectrum disentanglement~\cite{David2021}, visible-domain prior transfer \citep{mei2022escaping}, feature-aligned identity conditioning \citep{lin2024difftv} and multimodal conditional latent diffusion~\citep{MTVDiff2026}. Despite this progress, prior work has largely focused on improving the overall cross-spectral mapping, with less attention to how thermal cues of unequal reliability and identity information should be organized within the generative process. First, many conditional generators represent the thermal input as a homogeneous spatial condition, although facial structure, fine details, and sensor artifacts provide thermal-visible correspondence of unequal reliability. Processing them through a uniform conditioning pathway makes it difficult to preserve reliable structural cues while suppressing modality-specific or noisy responses, potentially weakening facial geometry or introducing unstable details. Second, spatial correspondence alone does not guarantee identity consistency. Existing identity losses and feature-alignment objectives provide useful training supervision, but they generally do not represent cross-spectral identity as an independent condition during denoising. Consequently, a generated face may appear visually plausible while deviating from the identity of the thermal input. Addressing both challenges is particularly difficult with limited paired thermal-visible face data, motivating the reuse of pretrained generative and identity priors.

% SOURCE-DERIVED (PARAPHRASED): DiT.
Pretrained DiT-based generators provide a strong visible-domain prior under limited paired data \citep{peebles2023dit}. We therefore construct a task-specific FLUX-T2V baseline that encodes the thermal image as an aligned condition-token stream while keeping the pretrained VAE and transformer frozen. This design focuses learning on cross-spectral control rather than visible-face synthesis. However, a single low-rank path must process all thermal cues, while identity remains embedded in the spatial condition. These limitations motivate frequency-specialized adaptation and a separate cross-spectral identity condition.

{\clubpenalty=10000
To examine the first limitation, we conduct a controlled cross-dataset frequency study, with the full protocol and numerical results reported in the supplementary material. As summarized in Figure~\ref{fig:freq-motivation}, pair-specific low-frequency structural correspondence is consistently observed across datasets, whereas high-frequency correspondence is less stable, more dataset-dependent, and concentrated around facial edges. Thermal evidence should therefore not be adapted uniformly. Although image-domain frequency is not equivalent to frequency in transformer hidden states, these observations motivate the testable hypothesis that separate low- and high-frequency update subspaces, combined through state-dependent routing, can better accommodate their different reliability.
\par}

\begin{figure}[h]
    \centering
    \includegraphics[width=\columnwidth]{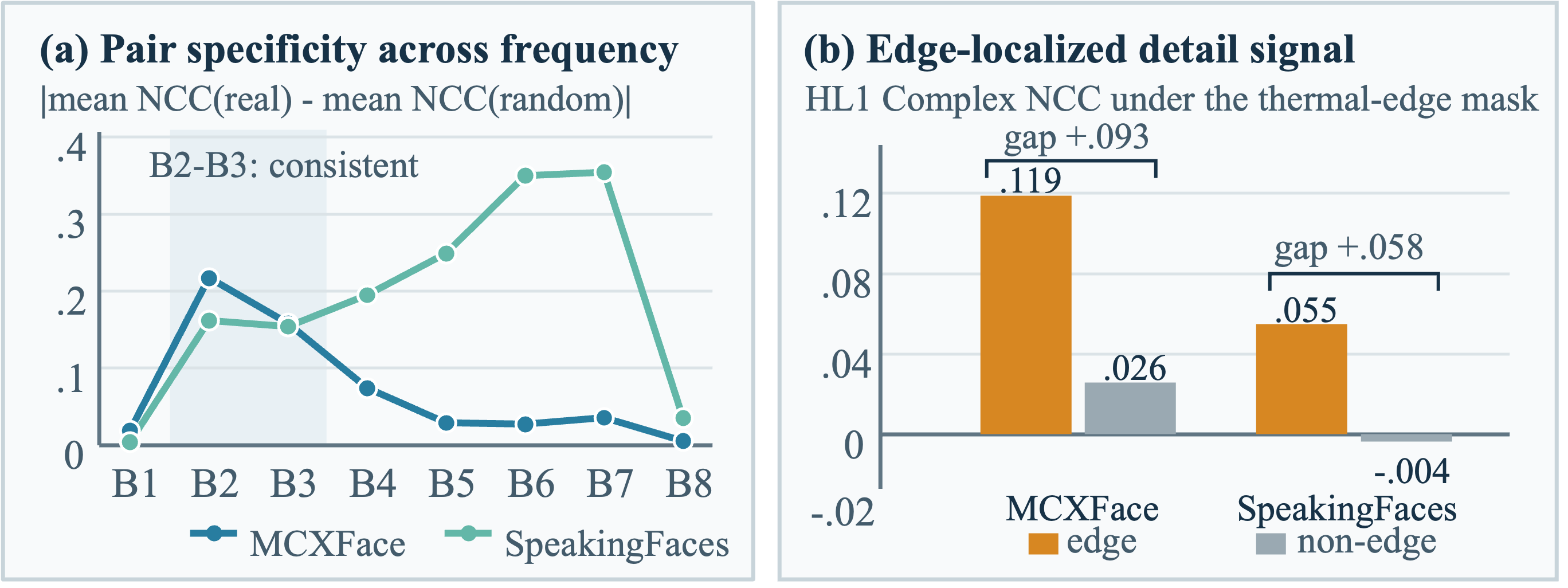}
    \caption{Frequency-domain motivation. (a) Absolute real--random mean Complex-NCC gaps across FFT bands. B2--B3 show pair-specific correspondence across datasets, whereas finer-frequency behavior varies. (b) HL1 Complex NCC at thermal-edge and non-edge locations, showing edge-concentrated transferable detail.}
    \label{fig:freq-motivation}
\end{figure}

We introduce \method{}, a parameter-efficient framework built on the baseline. Motivated by the frequency evidence above and the need for independent identity guidance, it organizes thermal information along two complementary axes: spectral reliability and semantic role. First, the thermal input is encoded as a spatial condition sequence by the baseline conditioning pathway, preserving its aligned facial layout. Frequency LoRA-MoE then decomposes intermediate token sequences into low- and high-frequency components and assigns them to separate low-rank experts, with state-adaptive routing controlling their contributions. In parallel, a pretrained cross-spectral identity module maps the same thermal face into a modality-shared identity space, and a lightweight resampler converts the resulting global representation into compact context tokens. The spatial and identity sequences remain distinct until multimodal attention, allowing local structure and global identity to guide generation without being collapsed into a single condition stream.

Our contributions are summarized as follows:

\begin{itemize}
    \item We establish a cross-dataset frequency profile of paired thermal-visible faces through real/random/self controls, sensor-noise suppression, and edge-conditioned analysis, providing an empirical rationale for frequency-selective adaptation.
    \item We introduce Frequency LoRA-MoE for parameter-efficient cross-spectral DiT adaptation. Separate low-rank subspaces model low-frequency structure and high-frequency detail, while learned and energy-based routing adjust their contributions without fine-tuning the generative backbone.
    \item We introduce Shared Cross-Spectral Identity Conditioning, which learns identity-discriminative factors jointly supported by thermal and visible faces rather than mapping thermal features toward a visible-specific representation. The resulting shared identity representation is resampled into compact tokens and injected through a condition branch separate from spatial thermal tokens.
\end{itemize}

\section{Related Work}

\subsection{T2V Face Translation}

T2V face translation maps a thermal observation to a visible-spectrum face. Early approaches learn this cross-spectral mapping using supervised synthesis networks. Multi-region models combine global facial prediction with locally specialized mappings to recover spatially varying details \citep{riggan2018thermal}, while attribute-preserving adversarial models use facial attributes to guide visible synthesis \citep{di2018attribute}. Later methods address data scarcity by exploiting abundant visible-face data to reduce reliance on paired thermal-visible samples \citep{mei2022escaping}. These methods establish facial structure and identity as central generation objectives, but generally learn visible-domain synthesis and cross-spectral transfer jointly within dedicated networks.

Diffusion models provide a probabilistic alternative to adversarial image translation. Palette formulates general image-to-image tasks as conditional denoising \citep{saharia2022palette}, whereas BBDM constructs a Brownian bridge directly between source and target domains \citep{li2023bbdm}. For thermal faces, T2V-DDPM adapts conditional diffusion to thermal-to-visible synthesis \citep{nair2023t2v}, and DiffTV combines latent diffusion with heterogeneous feature alignment and dual-stage conditioning for identity preservation \citep{lin2024difftv}. Recent approaches further incorporate facial attributes or thermal-depth-text cues to constrain visible appearance and identity \citep{cai2025multiattribute,xia2026mtvdiff}. Together, these studies show that diffusion models can combine generative priors with increasingly rich structural, semantic, and identity-related guidance for T2V synthesis.

Pretrained latent diffusion and diffusion transformers (DiTs) reduce the need to learn the visible-image distribution from limited paired data \citep{rombach2022ldm,peebles2023dit}. ControlNet attaches a trainable residual branch to a frozen diffusion model \citep{zhang2023controlnet}, while OminiControl encodes image conditions with the DiT's own VAE and processes condition and noisy-image tokens in a unified sequence using low-rank updates \citep{tan2025omini}. OminiControl2 focuses on reusing compressed condition features for efficient inference \citep{tan2025ominicontrol2}. These general-purpose control frameworks provide flexible mechanisms for image conditioning, but are not designed around the frequency-dependent reliability and cross-spectral identity requirements specific to T2V face generation.

\subsection{Cross-Spectral Identity}

Face-recognition models such as ArcFace learn discriminative angular geometry from visible-spectrum identities \citep{deng2019arcface}, but their embeddings are not inherently invariant to thermal-visible appearance changes. Heterogeneous face-recognition methods therefore introduce modality-specific transformations or shared embedding spaces to reduce the cross-spectral gap \citep{george2022prepended}. In generative T2V systems, identity is encouraged through recognition objectives, heterogeneous feature alignment, attribute guidance, or multimodal conditions \citep{lin2024difftv,cai2025multiattribute,xia2026mtvdiff}. DiffTV, in particular, explicitly injects aligned identity-related features through dual-stage conditioning \citep{lin2024difftv}. Across these designs, identity may serve as supervision, an aligned feature, or an explicit condition, but it is generally learned jointly with translation or integrated with other guidance streams. Comparatively less attention has been given to establishing a frozen cross-spectral identity representation before translation and exposing it as a condition stream distinct from spatial thermal features.

\section{Method}

\subsection{Problem Formulation and Baseline}

Let $I_T \in [0,1]^{3\times H\times W}$ be a thermal face and $I_V$ its paired visible target. A frozen variational autoencoder (VAE) encodes and packs them into token sequences
\begin{equation}
    c_T = \mathcal{P}(\mathcal{E}_{\mathrm{vae}}(I_T)), \qquad
    x_0 = \mathcal{P}(\mathcal{E}_{\mathrm{vae}}(I_V)),
\end{equation}
where $c_T,x_0\in\mathbb{R}^{N\times d_z}$ and $\mathcal{P}$ denotes FLUX's $2\times2$ latent packing. Following flow matching, we draw $x_1\sim\mathcal{N}(0,I)$ and $z\sim\mathcal{N}(0,1)$, set $t=\sigma(z)$, and construct
\begin{equation}
    x_t=(1-t)x_0+t x_1, \qquad v^{*}=x_1-x_0.
    \label{eq:flow-path}
\end{equation}
The baseline predicts $v_{\mathrm{base}}(x_t,c_T,t)$ while the FLUX backbone and VAE remain frozen.

The baseline represents the thermal image as an aligned spatial-token branch and learns the cross-spectral mapping through low-rank adaptation. We write token features row-wise. For $X\in\mathbb{R}^{N\times d_{\mathrm{in}}}$, a rank-$r_b$ thermal adapter modifies a selected linear projection as
\begin{equation}
    \mathcal{L}_{\mathrm{base}}(X)
    = XW^{\top}
    + \frac{\alpha_b}{r_b}(XA_b^{\top})B_b^{\top},
    \label{eq:baseline-lora}
\end{equation}
where $W\in\mathbb{R}^{d_{\mathrm{out}}\times d_{\mathrm{in}}}$ is frozen, $A_b\in\mathbb{R}^{r_b\times d_{\mathrm{in}}}$ is the down-projection, and $B_b\in\mathbb{R}^{d_{\mathrm{out}}\times r_b}$ is the up-projection. Thus, the trainable weight update is $\Delta W_b=(\alpha_b/r_b)B_bA_b$, while Eq.~\eqref{eq:baseline-lora} follows the row-token convention used by the implementation.
After baseline training, this adapter is loaded into \method{}, activated only for the spatial thermal branch, and kept frozen. It supplies an initial thermal-to-visible mapping upon which the frequency and identity components are optimized.

\begin{figure*}[t]
    \centering
    \includegraphics[width=\textwidth]{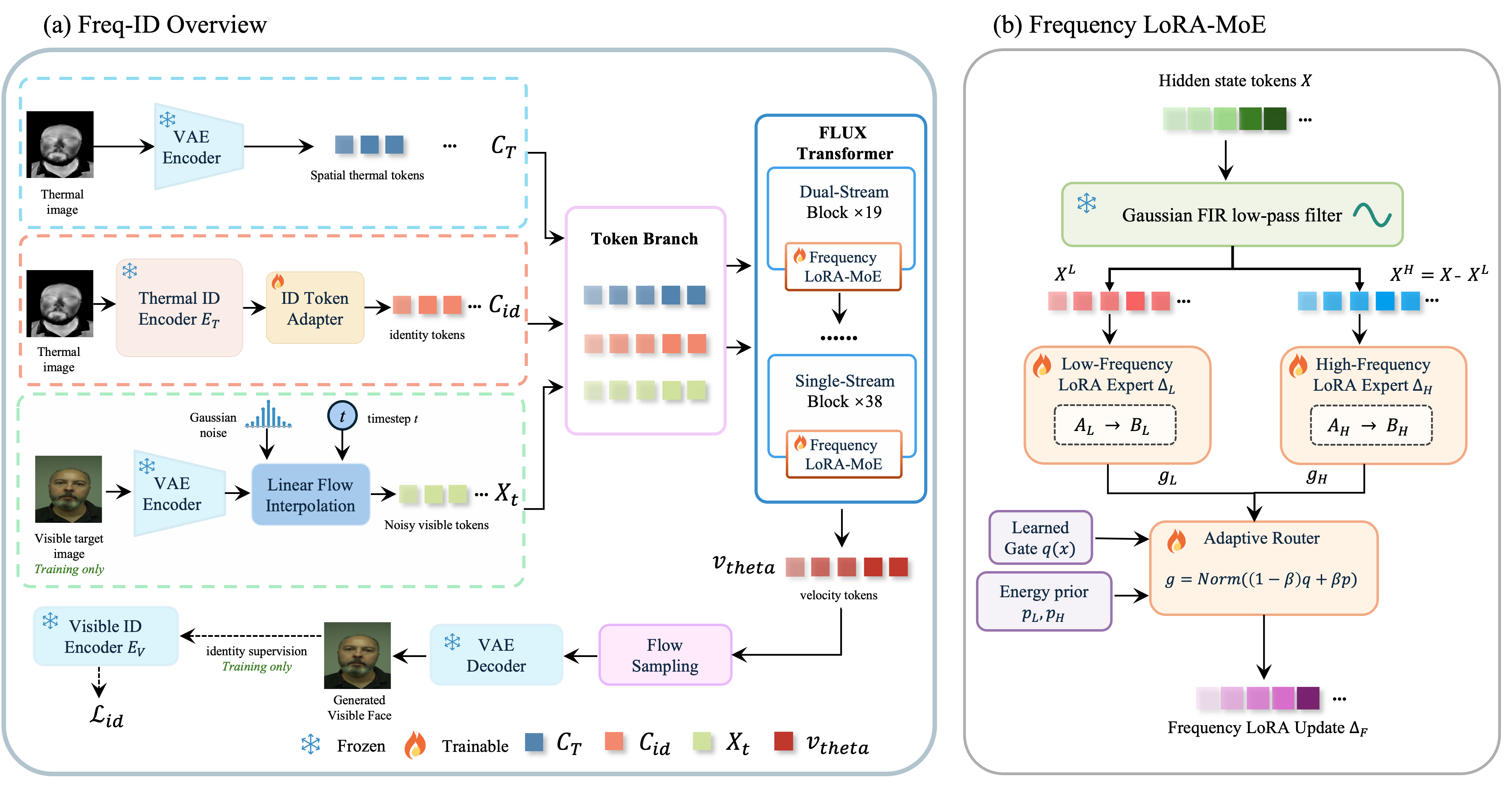}
    \caption{Overview of \method{}. The thermal input provides a spatial condition through the frozen FLUX-based T2V pathway and a global identity condition through the shared cross-spectral encoder and trainable resampler. Spatial, identity, and noisy-latent states remain separate until multimodal attention. Frequency LoRA-MoE decomposes each current hidden state into low- and high-frequency token components, applies specialized low-rank experts, and routes their updates back to the projection. Gray modules are frozen; orange and green modules are trainable.}
    \label{fig:framework}
\end{figure*}

Figure~\ref{fig:framework} summarizes the proposed framework. We model three input sequences: thermal identity tokens $C_{\mathrm{id}}$, noisy visible tokens $X_t$, and spatial thermal tokens $C_T$. The two thermal-derived branches serve different roles. $C_T$ preserves spatially aligned structure through the baseline conditioning path, whereas $C_{\mathrm{id}}$ provides a compact global identity cue. Branch states are embedded independently and then interact through FLUX multimodal attention:
\begin{equation}
    [C_{\mathrm{id}}^{\ell+1},X_t^{\ell+1},C_T^{\ell+1}]
    =\mathcal{B}_{\ell}([C_{\mathrm{id}}^{\ell},
      X_t^{\ell},C_T^{\ell}]).
\end{equation}
Thus, the source information is shared across blocks, but the injected representation is not constant: every block updates each branch and recomputes its frequency routing from the current hidden states.

\subsection{Frequency LoRA-MoE}

Low-rank adaptation and expert routing provide complementary mechanisms for
parameter-efficient residual learning and conditional specialization
\citep{hu2022lora,shazeer2017moe}. We combine these principles with a
lightweight token-frequency decomposition tailored to the measured reliability
of thermal-visible correspondence.

\paragraph{Token-frequency decomposition.}
Consider the input $X\in\mathbb{R}^{B\times N\times d_{\mathrm{in}}}$ to a patched attention projection. We apply a fixed depthwise Gaussian finite-impulse-response kernel $K$ along the token dimension:
\begin{equation}
    X^{L}=K*X, \qquad X^{H}=X-X^{L}.
    \label{eq:freq-split}
\end{equation}
The operation is performed independently for each feature channel and branch. Because packed image tokens follow raster order, it captures low- versus high-frequency variation along the token sequence with negligible parameters. We use the term \emph{token frequency} to distinguish this operation from an exact two-dimensional Fourier decomposition.

\paragraph{Frequency-specialized LoRA experts.}
Two LoRA experts process the low-pass and residual high-pass components, respectively:
\begin{equation}
    \Delta_i(X^i)=\frac{\alpha_i}{r_i}
    (X^iA_i^{\top})B_i^{\top},
    \qquad i\in\{L,H\},
    \label{eq:experts}
\end{equation}
where $A_i\in\mathbb{R}^{r_i\times d_{\mathrm{in}}}$ and $B_i\in\mathbb{R}^{d_{\mathrm{out}}\times r_i}$. The low-frequency expert models slowly varying structure, whereas the high-frequency expert models rapidly varying detail. Their independent low-rank subspaces prevent a single adapter from using the same update for both regimes.

\paragraph{Learned-spectral routing.}
The router combines a learned token-wise distribution with a sample-level component-energy prior. For each token, the learned distribution is
\begin{equation}
    q(X)=\operatorname{softmax}(XG^{\top}/\tau),
    \qquad G\in\mathbb{R}^{2\times d_{\mathrm{in}}},
\end{equation}
where the softmax is taken over the two experts and $\tau$ is the temperature. From the decomposed components, we estimate the per-sample energies
\begin{equation}
\begin{aligned}
    e_i&=\frac{1}{Nd_{\mathrm{in}}}\lVert X^i\rVert_F^2, \\
    p_H&=\frac{e_H}{e_L+e_H+\epsilon},
    &p_L&=1-p_H.
\end{aligned}
\end{equation}
where $i\in\{L,H\}$. The prior $p=[p_L,p_H]$ is broadcast over the token dimension. Let $\beta=\sigma(\rho)$ be a learned scalar for each projection. We blend the learned distribution and the broadcast energy prior, then normalize over the two experts:
\begin{equation}
\begin{aligned}
    \widetilde{g}_{n,i}&=(1-\beta)q_{n,i}+\beta p_i, \\
    g_{n,i}&=\frac{\widetilde{g}_{n,i}}
    {\sum_{j\in\{L,H\}}\widetilde{g}_{n,j}+\epsilon}.
\end{aligned}
    \label{eq:gate}
\end{equation}
Although $t$ is not supplied directly to $G$, routing is recomputed at every denoising step from the evolving hidden state and is therefore state-adaptive. With two experts and $k=2$, both experts remain active and the gate acts as a soft mixture rather than sparse dispatch.

At each patched projection, the two gated expert outputs are summed and added as a scaled residual to the projection output. The frozen baseline LoRA remains active only on the spatial thermal branch. Frequency adapters are installed in Q/K/V projections of all 19 dual-stream and 38 single-stream FLUX blocks, and in the output projection of each dual-stream attention block, for 190 patched linear layers in total.

To promote specialization, we penalize high-pass leakage from the low expert and low-pass leakage from the high expert:
\begin{equation}
    \mathcal{L}_{\mathrm{freq}}=
    \sum_{\ell}\left(
    \frac{\lVert \Delta_L^{\ell}-K*\Delta_L^{\ell}\rVert_F^2}
         {|\Delta_L^{\ell}|}
    +\frac{\lVert K*\Delta_H^{\ell}\rVert_F^2}
          {|\Delta_H^{\ell}|}\right),
    \label{eq:freq-reg}
\end{equation}
where $|\Delta_i^{\ell}|$ denotes the number of elements in the corresponding expert output, matching the mean-squared reduction used in training.

\subsection{Cross-Spectral Identity Conditioning}

\paragraph{Learning a shared cross-spectral identity space.}
Our objective is to retain identity factors jointly supported by thermal and visible observations while suppressing modality-specific appearance. We use modality-specific shallow stems $E_T$ and $E_V$ to absorb sensor-dependent statistics, followed by a shared IResNet body $S$ and a common projector $P_{\mathrm{id}}$. The network is initialized from an IResNet-100 pretrained with ArcFace on Glint360K, but the visible branch is not a frozen teacher. Both modalities jointly optimize the shared identity representation:
\begin{equation}
    f_m=\operatorname{norm}\!\left(
    P_{\mathrm{id}}(S(E_m(I_m)))\right),
    \qquad m\in\{T,V\}.
\end{equation}
The modality-specific stems and shared body are optimized at different learning rates, allowing shallow layers to adapt to spectral statistics without rapidly overwriting the pretrained identity geometry. A shared angular-margin classifier imposes common identity decision boundaries on both modalities:
\begin{equation}
    \mathcal{L}_{\mathrm{cls}}=
    \frac{1}{2}\left(\mathcal{L}^{T}_{\mathrm{arc}}
    +\mathcal{L}^{V}_{\mathrm{arc}}\right).
\end{equation}
To directly preserve identity evidence shared across spectra, we maximize the similarity of all same-identity cross-modal samples,
\begin{equation}
    \mathcal{L}_{\mathrm{pair}}=
    1-\frac{1}{|\mathcal{P}|}
    \sum_{(i,j)\in\mathcal{P}} f_{T,i}^{\top}f_{V,j},
\end{equation}
where $\mathcal{P}=\{(i,j):y_{T,i}=y_{V,j}\}$. A symmetric multi-positive InfoNCE objective uses a cross-modal feature queue to attract same-identity samples and separate different identities in both retrieval directions. Identity prototypes $\bar f_{T,k}$ and $\bar f_{V,k}$ further define relation matrices $R_T$ and $R_V$, with entries $R_{ab}=\bar f_a^{\top}\bar f_b$. Their off-diagonal discrepancy is regularized by
\begin{equation}
    \mathcal{L}_{\mathrm{rel}}=
    \operatorname{SmoothL1}(R_T,R_V).
\end{equation}
Finally, a modality discriminator $D$ receives the embeddings through a gradient-reversal layer (GRL):
\begin{equation}
    \mathcal{L}_{\mathrm{dom}}=\operatorname{CE}\!\left(
    D(\operatorname{GRL}_{\lambda_d}([f_T;f_V])),m\right).
\end{equation}
The discriminator learns to identify the input spectrum, whereas the reversed gradient discourages the identity encoder from retaining modality-predictive directions. The complete identity-pretraining objective is
\begin{equation}
\begin{aligned}
    \mathcal{L}_{\mathrm{pre}}={}&
    \lambda_a\mathcal{L}_{\mathrm{cls}}
    +\lambda_p\mathcal{L}_{\mathrm{pair}} \\
    &+\lambda_c\mathcal{L}_{\mathrm{nce}}
    +\lambda_r\mathcal{L}_{\mathrm{rel}}
    +\mathcal{L}_{\mathrm{dom}}.
\end{aligned}
\end{equation}
This formulation does not regress thermal embeddings toward a fixed visible target. Instead, modality-specific stems absorb spectrum-dependent observations, while shared parameters, symmetric cross-modal objectives, and domain confusion encourage the common space to retain identity-discriminative factors supported by both spectra. After pretraining, the entire identity encoder is frozen. The thermal branch provides the identity condition, and the visible branch embeds generated faces for identity supervision during translation training.

\paragraph{Identity-token resampling.}
At translation time, only the thermal embedding is available. A trainable two-layer transformer resampler uses $M$ learned queries to map $f_T$ into the FLUX context dimension:
\begin{equation}
    C_{\mathrm{id}}=\mathcal{R}(f_T;Q)\in\mathbb{R}^{M\times d}.
\end{equation}
We use $M=4$. These tokens bypass the standard context embedder, receive distinct positional identifiers, and enter attention as an independent context branch. The identity representation is therefore not merged into the spatial thermal sequence or used only as a training loss. Instead, compact global identity tokens and spatially aligned thermal tokens remain structurally distinct until multimodal attention determines their interaction.

\subsection{Training Objectives}

The primary flow-matching objective is
\begin{equation}
    \mathcal{L}_{\mathrm{flow}}
    =\lVert v_\theta(x_t,c_T,C_{\mathrm{id}},t)-v^{*}\rVert_2^2.
\end{equation}
For sufficiently low noise, the predicted clean latent is
\begin{equation}
    \hat{x}_0=x_t-t v_\theta(x_t,c_T,C_{\mathrm{id}},t).
\end{equation}
After VAE decoding, the frozen visible identity branch extracts $f_{\mathrm{gen}}$. We define
\begin{equation}
    \mathcal{L}_{\mathrm{id}}=
    \mathbf{1}[t\leq t_{\max}](1-t)^{\gamma}
    \left(1-\cos(f_{\mathrm{gen}},f_T)\right).
    \label{eq:id-loss}
\end{equation}
The complete objective is
\begin{equation}
    \mathcal{L}=\mathcal{L}_{\mathrm{flow}}
    +\lambda_f\mathcal{L}_{\mathrm{freq}}
    +\lambda_{\mathrm{id}}\mathcal{L}_{\mathrm{id}}.
    \label{eq:total-loss}
\end{equation}
The current configuration sets $\lambda_f=10^{-3}$, $\lambda_{\mathrm{id}}=0.05$, $t_{\max}=0.35$, and $\gamma=2$. The trainable parameters comprise the frequency LoRA experts, learned gates and blend coefficients, and identity resampler; all generative backbone parameters, the baseline spatial-conditioning adapter, and the cross-spectral identity encoder are frozen.

\section{Experiments}

We evaluate \method{} through translation quality, biometric identity
preservation, qualitative comparison, and controlled component ablations. The
cross-dataset frequency analysis that motivates the architecture is reported in
the supplementary material.

\begin{table*}[!t]
\centering
{\small
\setlength{\tabcolsep}{3pt}
\begin{tabular*}{\textwidth}{@{\extracolsep{\fill}}lcccccccc@{}}
\toprule
& \multicolumn{4}{c}{MCXFace} &
\multicolumn{4}{c}{SpeakingFaces} \\
\cmidrule(lr){2-5}\cmidrule(l){6-9}
Method & FID $\downarrow$ & LPIPS $\downarrow$ & SSIM $\uparrow$ &
PSNR $\uparrow$ & FID $\downarrow$ & LPIPS $\downarrow$ & SSIM $\uparrow$ &
PSNR $\uparrow$ \\
\midrule
Axial-GAN & 129.62 & 0.2131 & 0.6441 & 17.82 &
46.69 & 0.1729 & 0.6673 & 20.45 \\
BBDM & 127.06 & 0.1926 & 0.7629 & 20.50 &
41.40 & 0.2112 & 0.7161 & 26.51 \\
AT-DDPM & 116.02 & 0.2621 & 0.6402 & 17.39 &
71.34 & 0.4063 & 0.5554 & 8.75 \\
T2V-DDPM & 120.65 & 0.2445 & 0.6740 & 18.37 &
36.33 & 0.2705 & 0.6832 & 15.53 \\
DiffTV & -- & -- & -- & -- &
34.02 & 0.1650 & 0.7504 & \textbf{29.45} \\
Uni-ControlNet & 97.01 & 0.2151 & 0.7181 & 20.01 &
29.20 & 0.3017 & 0.7042 & 19.81 \\
MTVDiff (thermal-only) & 93.96 & 0.2001 & 0.7367 & 20.45 &
\textbf{20.14} & \textbf{0.1320} & 0.7836 & 20.91 \\
DiffV2IR (thermal-only) & 89.71 & 0.2220 & 0.6718 & 19.56 &
22.13 & 0.2007 & 0.7609 & 20.93 \\
\method{} & \textbf{74.86} & \textbf{0.1626} & \textbf{0.7647} &
\textbf{21.25} & 21.74 & 0.1637 &
\textbf{0.7969} & 21.96 \\
\bottomrule
\end{tabular*}
}
\caption{Reported thermal-to-visible image-quality results on MCXFace and
SpeakingFaces. Published baselines follow their respective evaluation
protocols; the best reported value for each dataset and metric is shown in
bold. Dashes indicate unavailable results.}
\label{tab:main-quality}
\end{table*}

\subsection{Experimental Setup}

\paragraph{Datasets.}
MCXFace provides spatially registered observations from 51 identities acquired
over three sessions under multiple illumination settings
\citep{george2022prepended}. Following the T2V preprocessing protocol, we form
6,120 thermal-visible pairs at $256\times256$ resolution. The identity-disjoint
partition assigns 41 subjects (4,920 pairs) to training and the remaining 10
subjects (1,200 pairs) to testing. Within the training identities, cross-spectral
identity pretraining uses 33 subjects for optimization and 8 for validation.

SpeakingFaces records synchronized thermal and visible video from 142
participants \citep{abdrakhmanova2021speakingfaces}. After removing invalid or
poorly aligned frames, our protocol retains 7,700 paired face samples spanning
nine viewpoint configurations. We use 5,400 pairs from 100 identities for
training and 2,300 pairs from 42 disjoint identities for testing. No test
identity is used to optimize either the translation model or the
cross-spectral identity encoder.

\paragraph{Implementation details.}
We use FLUX.1-dev as the frozen generative backbone. The Frequency
LoRA-MoE is installed in the attention projections of all 19 dual-stream and 38
single-stream transformer blocks. Unless otherwise stated, the low- and
high-frequency experts use rank 8, a length-9 FIR kernel, temperature 5,
residual scale 0.1, and soft two-expert routing. The identity resampler produces
four 3,072-dimensional context tokens. Translation training runs for 8,000
optimizer updates using bfloat16, an effective batch size of 8, gradient
clipping at 0.5, and Prodigy with weight decay 0.01. Inputs and targets are
directly resized to $256\times256$. We sample each image with 28 denoising
steps, guidance scale 1.8, image-guidance scale 1.0, and seed 42.

The cross-spectral identity encoder uses an ArcFace IResNet-100 initialized from
Glint360K. Thermal and visible images pass through modality-specific shallow
stems and a shared identity body; both modalities are jointly optimized for
30,000 steps, with a lower learning rate applied to the shared pretrained
layers. Training combines angular-margin classification, paired alignment,
cross-modal InfoNCE, relation preservation, and domain confusion. The complete
identity encoder is frozen before translation training.

\paragraph{Metrics and baselines.}
Translation quality is measured with FID, LPIPS, PSNR, and SSIM. For biometric
evaluation, ArcFace embeddings are used to compute Rank-1 identification and
verification rates at false-acceptance rates of $1\%$ and $0.1\%$.
Embedding-based identity scores are computed on images for which the face
detector returns a valid face, and the number of valid detections is recorded
separately. Image
metrics are computed at $256\times256$ with PyIQA using matched generated and
reference images. We compare against available adversarial and diffusion T2V
systems under their reported dataset protocols; controlled ablations evaluate
the baseline, frequency-only, identity-only, and complete variants
on a common test manifest.

\subsection{Main Translation Results}

As shown in Table~\ref{tab:main-quality}, \method{} achieves the best reported
MCXFace results across all four image-quality metrics. In particular, it improves
SSIM from the previous best of 0.7629 to 0.7647, indicating stronger preservation
of cross-spectrally supported facial structure. This structural improvement is
accompanied by consistent gains in distributional, perceptual, and pixel-level
quality: FID decreases from 89.71 to 74.86, LPIPS from 0.1926 to 0.1626, and
PSNR increases from 20.50 to 21.25 dB. These results show that \method{} improves
structural consistency without sacrificing visible-domain synthesis quality.

On SpeakingFaces, \method{} leads in SSIM and the identity measures in
Table~\ref{tab:identity-quality}, but trails thermal-only MTVDiff in FID and
LPIPS and DiffTV in PSNR. This profile matches our emphasis on
cross-spectrally supported structure and identity rather than pixel-level
recovery of visible-only appearance, indicating a metric trade-off rather than
uniform superiority.

\begin{figure*}[!t]
    \centering
    \includegraphics[width=\textwidth]{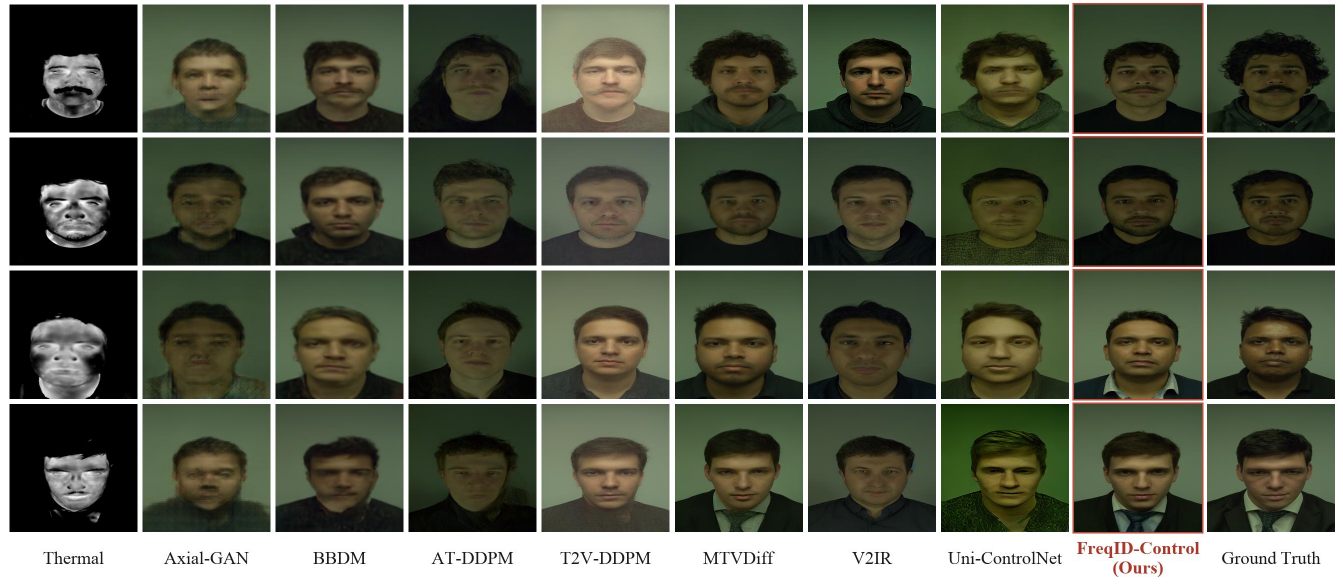}
    \caption{Qualitative comparison on four MCXFace test samples. DiffV2IR uses its
    thermal-only setting; the FreqID-Control column is outlined, and Ground
    Truth is shown last.}
    \label{fig:model_comparison}
\end{figure*}

Figure~\ref{fig:model_comparison} compares MCXFace results with varied facial
structure, hairstyle, and facial hair. Earlier methods frequently
blur facial boundaries or alter identity-related geometry. In comparison,
\method{} more consistently preserves the principal structure of the thermal
observation while producing a coherent visible-spectrum face.

\subsection{Ablation Study}

\begin{center}
\begin{minipage}{\columnwidth}
\centering
{\small
\setlength{\tabcolsep}{1.8pt}
\begin{tabular*}{\columnwidth}{@{\extracolsep{\fill}}ccccccc@{}}
\toprule
\shortstack{Frequency\\LoRA-MoE} & \shortstack{Shared\\ID} &
FID $\downarrow$ & LPIPS $\downarrow$ & SSIM $\uparrow$ &
PSNR $\uparrow$ & Rank-1 $\uparrow$ \\
\midrule
-- & -- & 77.95 & 0.1990 & 0.7060 & 18.63 & 0.8725 \\
$\checkmark$ & -- & 75.13 & 0.1698 & 0.7494 & 21.12 & 0.9471 \\
-- & $\checkmark$ & 74.84 & 0.1779 & 0.7362 & 20.86 & 0.9271 \\
$\checkmark$ & $\checkmark$ & 74.86 & \textbf{0.1626} & \textbf{0.7647} &
\textbf{21.25} & \textbf{0.9682} \\
\bottomrule
\end{tabular*}
}
\captionof{table}{Component ablation on MCXFace.}
\label{tab:ablation-design}
\end{minipage}
\end{center}

\paragraph{Component effects.}
All variants in Table~\ref{tab:ablation-design} share the same baseline and test
manifest. Frequency LoRA-MoE improves every metric, including a 2.49 dB PSNR
gain and a Rank-1 increase from 0.8725 to 0.9471, supporting its role in adapting
cross-spectral structure and detail. Identity conditioning alone lowers FID to
74.84 and raises Rank-1 to 0.9271, showing that shared identity tokens add a
global modality-compatible cue beyond spatial conditioning. Combining both
components yields the best LPIPS, SSIM, PSNR, and Rank-1 (0.9682), while its FID
is effectively tied with identity conditioning alone. This result supports
their complementary roles in frequency-aware structural adaptation and global
cross-spectral identity guidance.

\begin{center}
\begin{minipage}{\columnwidth}
\centering
\setlength{\tabcolsep}{0.8pt}
\begin{tabular}{ccccc}
{\scriptsize Baseline} & {\scriptsize Frequency} & {\scriptsize Shared ID} &
{\scriptsize Full} & {\scriptsize GT} \\
\includegraphics[width=0.185\columnwidth]{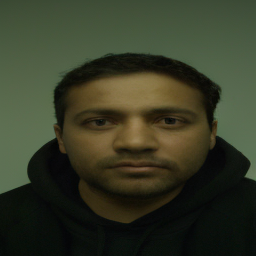} &
\includegraphics[width=0.185\columnwidth]{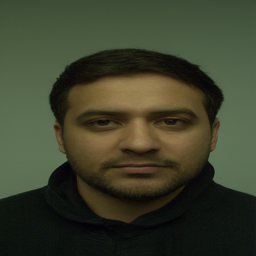} &
\includegraphics[width=0.185\columnwidth]{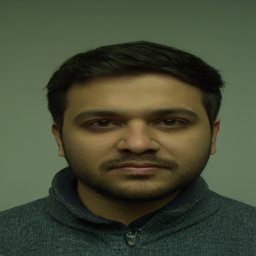} &
\includegraphics[width=0.185\columnwidth]{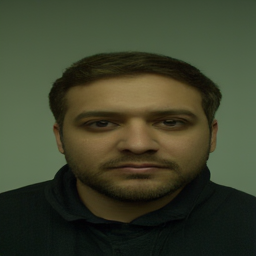} &
\includegraphics[width=0.185\columnwidth]{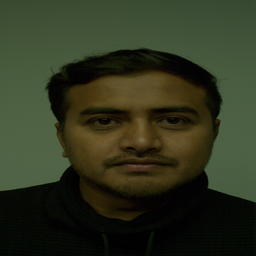} \\
\includegraphics[width=0.185\columnwidth]{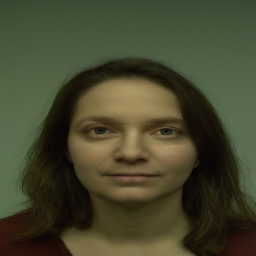} &
\includegraphics[width=0.185\columnwidth]{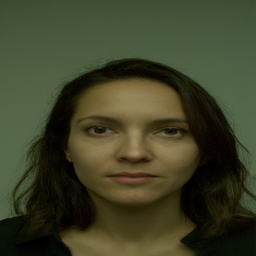} &
\includegraphics[width=0.185\columnwidth]{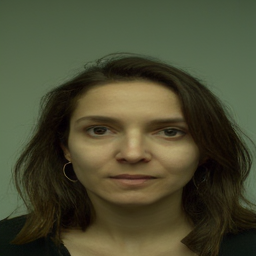} &
\includegraphics[width=0.185\columnwidth]{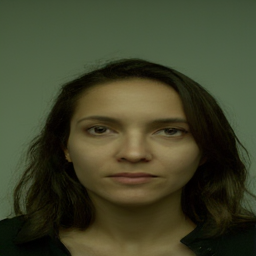} &
\includegraphics[width=0.185\columnwidth]{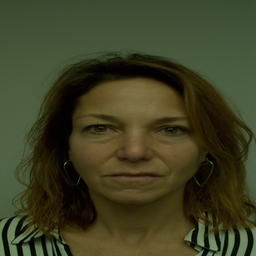}
\end{tabular}
\captionof{figure}{Qualitative component ablation on two MCXFace samples. The
full model combines Frequency LoRA-MoE and shared identity conditioning.}
\label{fig:component-ablation}
\end{minipage}
\end{center}

\begin{table*}[!t]
\centering
{\small
\setlength{\tabcolsep}{5pt}
\begin{tabular*}{\textwidth}{@{\extracolsep{\fill}}lcccccc@{}}
\toprule
& \multicolumn{3}{c}{MCXFace} &
\multicolumn{3}{c}{SpeakingFaces} \\
\cmidrule(lr){2-4}\cmidrule(l){5-7}
Method & Rank-1 & VR@$1\%$ & VR@$0.1\%$ &
Rank-1 & VR@$1\%$ & VR@$0.1\%$ \\
\midrule
Axial-GAN & 0.6815 & 0.0833 & 0.0033 & 0.2357 & 0.1606 & 0.0402 \\
BBDM & 0.7571 & 0.0960 & 0.0037 & 0.8659 & 0.5933 & 0.1257 \\
AT-DDPM & 0.5664 & 0.0704 & 0.0122 & 0.4169 & 0.3422 & 0.0782 \\
T2V-DDPM & 0.6815 & 0.1154 & 0.0256 & 0.4940 & 0.3357 & 0.0816 \\
DiffTV & -- & -- & -- & 0.6885 & 0.0754 & 0.0159 \\
DiffV2IR & 0.7834 & 0.1090 & 0.0192 & 0.5159 & 0.3660 & 0.0679 \\
Uni-ControlNet & 0.3758 & 0.0630 & 0.0126 & 0.8619 & 0.6987 &
0.4555 \\
MTVDiff (multimodal) & 0.8726 & \textbf{0.2774} & \textbf{0.0258} &
0.9376 & 0.8011 & 0.3131 \\
\method{} & \textbf{0.9682} & 0.1338 & 0.0127 & \textbf{0.9688} &
\textbf{0.9000} & \textbf{0.6353} \\
\bottomrule
\end{tabular*}
}
\caption{Cross-spectral face-recognition results on MCXFace and SpeakingFaces.
Published baselines follow the evaluation compiled by MTVDiff
\citep{xia2026mtvdiff}; its row uses thermal--depth--text input, whereas
\method{} uses only thermal imagery. Results from different recognition
protocols are reference comparisons. Best values are bold, and dashes denote
unavailable results.}
\label{tab:identity-quality}
\end{table*}

\subsection{Identity Preservation Results}

Table~\ref{tab:identity-quality} shows that \method{} reaches Rank-1 scores of
0.9682 on MCXFace and 0.9688 on SpeakingFaces. The controlled ablation in
Table~\ref{tab:ablation-design} attributes this gain partly to Shared ID, which
raises MCXFace Rank-1 from 0.8725 to 0.9271; its further increase to 0.9682 with
Frequency LoRA-MoE demonstrates their complementary identity and structural
guidance. On SpeakingFaces, the full model also obtains the highest reported
Rank-1 and verification rates under the reported protocol. On MCXFace, however,
MTVDiff retains higher VR@$1\%$ and VR@$0.1\%$ despite its lower Rank-1. These
results show a consistent advantage in Rank-1 identification on both datasets,
but not uniform superiority at all verification operating points. Because the
published rows follow different evaluation protocols, comparisons beyond our
controlled ablation should be interpreted as reference results.

\FloatBarrier

\section{Conclusion}

We presented \method{}, a parameter-efficient T2V framework that separates
spatial thermal control from cross-spectral identity guidance. Frequency
LoRA-MoE adapts low-frequency structure and selectively reliable high-frequency
detail, while the shared identity pathway extracts identity-discriminative
factors supported by both spectra. Controlled MCXFace ablations show that the
two modules improve the FLUX-T2V baseline and are complementary. Across
MCXFace and SpeakingFaces, \method{} achieves strong structural and identity
results, but not every image-quality metric or verification operating point.
This reflects our objective of preserving cross-spectrally supported geometry
and identity; visible-only attributes absent from thermal observations are
generated plausibly rather than recovered.

\bibliography{aaai2027}

\end{document}